\documentclass{article}

\usepackage{PRIMEarxiv}

\usepackage[utf8]{inputenc} 
\usepackage[T1]{fontenc}    
\usepackage{hyperref}       
\usepackage{url}            
\usepackage{booktabs}       
\usepackage{amsfonts}       
\usepackage{nicefrac}       
\usepackage{microtype}      
\usepackage{lipsum}
\usepackage{fancyhdr}       
\usepackage{graphicx}       
\usepackage{caption}
\usepackage{subcaption}
\usepackage{tikz}
\usepackage{xspace}
\usepackage{orcidlink}
\usetikzlibrary{fit}
\usetikzlibrary{positioning}
\usetikzlibrary{shapes.geometric}

\newcommand{\pendigits}{\textsc{pendigits}\xspace}
\newcommand{\digits}{\textsc{digits}\xspace}
\newcommand{\gas}{\textsc{gas sensor}\xspace}
\newcommand{\iris}{\textsc{iris}\xspace}
\newcommand{\cover}{\textsc{cover type}\xspace}
\newcommand{\glass}{\textsc{glass}\xspace}
\newcommand{\wine}{\textsc{wine}\xspace}
\newcommand{\mice}{\textsc{mice}\xspace}
\newcommand{\tcga}{\textsc{tcga}\xspace}

\definecolor{caribbeangreen}{rgb}{0.0, 0.8, 0.6}
\definecolor{darkseagreen}{rgb}{0.56, 0.74, 0.56}

\graphicspath{{media/}}     

\title{Revisiting Silhouette Aggregation
}

\author{
John Pavlopoulos\\
Department of Informatics \\
Athens University of Economics and Business \\
Archimedes/Athena RC, Greece\\ 
\texttt{annis@aueb.gr} \\
\And
Georgios Vardakas\\
Dept. of Computer Science \& Engineering\\
University of Ioannina\\
GR 45110, Ioannina, Greece\\
\texttt{g.vardakas@uoi.gr}\\
\And
Aristidis Likas\\
Dept. of Computer Science \& Engineering\\
University of Ioannina\\
GR 45110, Ioannina, Greece\\
\texttt{arly@cs.uoi.gr}\\
}

\begin{document}
\maketitle

\begin{abstract}
Silhouette coefficient is an established internal clustering evaluation measure that produces a score per data point, assessing the quality of its clustering assignment. To assess the quality of the clustering of the whole dataset, the scores of all the points in the dataset are typically (micro) averaged into a single value. An alternative path, however, that is rarely employed, is to average first at the cluster level and then (macro) average across clusters. As we illustrate in this work with a synthetic example, the typical micro-averaging strategy is sensitive to cluster imbalance while the overlooked macro-averaging strategy is far more robust. By investigating macro-Silhouette further, we find that uniform sub-sampling, the only available strategy in existing libraries, harms the measure's robustness against imbalance. We address this issue by proposing a per-cluster sampling method. An experimental study on eight real-world datasets is then used to analyse both coefficients in two clustering tasks.
\end{abstract}


\section{Introduction}
The silhouette coefficient~\cite{rousseeuw1987silhouettes} serves as a widely used measure for assessing the quality of clustering assignments of individual data points. It produces scores on a scale from $-1$ to $1$ reflecting poor to excellent assignments, respectively. In real world applications, where it is widely accepted~\cite{layton2013evaluating,bafna2016document,tambunan2020electrical}, it is common practice to average these scores to derive a single (micro-averaged) value for the entire dataset.
This is the originally proposed aggregation strategy~\cite{rousseeuw1987silhouettes} and the only implementation in the popular \href{https://scikit-learn.org/stable/modules/generated/sklearn.metrics.silhouette_score.html}{\sc scikit-learn} machine learning library in Python. An alternative aggregation strategy, however, is to average the Silhouette scores per cluster and then (macro) average across clusters, but our exploration of the related literature shows that this is a strategy that is rarely used in published studies. This is an alarming finding, because micro-averaging, e.g., in a classification context, is known to be sensitive to class imbalance \cite{gaudreault2024empirical,suhaimi2022comparative}.

In this work, we focus on the effect of micro-averaging to a very well known internal cluster validity index, addressing the following research question \textbf{RQ1:} \emph{Is micro-averaging, which is the typical strategy to aggregate Silhouette scores in cluster analysis, sensitive to cluster imbalance?} We answer this question using synthetic data, showing that micro-averaging Silhouette can produce misleading results for clustering solutions with imbalanced clusters. By contrast, we show that macro-averaging, which is rarely used in literature, is considerably more robust to this issue, because it assigns equal weight per cluster while disregarding its size. 

In cluster analysis, Silhouette scores are often subsampled before being aggregated to yield a single score for a large dataset. This is a particularly useful step for computationally expensive tasks, as for example when assessing clustering solutions for a varying number of clusters to select the optimal \cite{schubert2023stop}. By evaluating existing libraries in Python and R, we observe that only uniform sampling is implemented, which makes us focus on a second research question \textbf{RQ2:} \emph{Is uniform sampling suitable when macro-averaging or is its robustness against cluster-imbalance put at stake?} The answer is the latter. Theoretically, in an extremely imbalanced dataset, the smallest cluster could even disappear when sub-sampling uniformly, which would exclude one (equal) factor from the macro-average. We address this issue, by proposing a novel per-cluster sampling strategy, which we show that it best suits macro-averaging of Silhouette scores.   

Overall, the contributions of this study can be summarised in the following three aspects:
\begin{itemize}
    \item We compare two aggregation strategies that can be used to compute a Silhouette score for a dataset, showing that the typical micro-averaging strategy is problematic for imbalanced datasets.
    \item We introduce a per-cluster sampling strategy, which should be the one used along with macro-averaging. 
    \item We quantify the sensitivity of micro-averaged Silhouette on imbalanced synthetic data, and we analyse two real-world imbalanced datasets on which the macro average should be preferred.
\end{itemize}

The remainder of this study comprises the related work (\S\ref{sec:literature}), a description of the Silhouette Coefficient (\S\ref{sec:silhouette}) and the aggregating strategies (\S\ref{sec:methods}), followed by an investigation on synthetic data (\S\ref{sec:synthetic}), an experimental study on real-word data (\S\ref{sec:experiments}), and closing with remarks and future directions. Our code is publicly available in \href{https://github.com/ipavlopoulos/revisiting-silhouette-aggregation}{https://github.com/ipavlopoulos/revisiting-silhouette-aggregation}.

\section{Related Work}\label{sec:literature}

\paragraph{The typical approach} The vast majority of published studies employ micro-averaging to report the Silhouette Coefficient. In \cite{azimi2017novel,dudek2020silhouette,unlu2019estimating}, the authors focus on the number of clusters estimation problem. In \cite{batool2021clustering}, they use the term Average Silhouette Width to refer to the micro-averaged Silhouette score. The authors of \cite{shahapure2020cluster} explicitly report the implementation of \textsc{scikit-learn} that employs the micro-averaging strategy. The authors of \cite{kang2016recursive}, proposed a clustering algorithm that divides recursively the clustered dataset based on the maximization of the silhouette index. Similarly with the rest studies, they used the micro-average strategy, which they called as summation of silhouettes.

\paragraph{Exceptions to the rule} 
In \cite{vrezankova2018different}, the author observes that SPSS and R employ different implementations of the Silhouette score. They note that the latter is using micro-averaging and is the correct out of the two. We observe, however, 
that other libraries (packages) do not necessarily follow this paradigm. ClusterCrit,\footnote{\url{https://cran.r-project.org/web/packages/clusterCrit/vignettes/clusterCrit.pdf}} for example, compute cluster mean silhouette scores that they average to yield the final index, but this is in fact a macro-averaged score.
Without any study in published literature to assess the two strategies, we argue that this is considerably problematic, because, as we show (\S\ref{sec:experiments}), results reported with the two strategies on the same data may not be comparable. Notably, we could only detect just one study using (and explicitly stating) the macro-averaging implementation \cite{brun2007model}. 

\paragraph{Filling the gap} Micro-averaging is the typical and widely-used approach when aggregating Silhouette, with macro-averaging being considerably overlooked in literature of cluster analysis. Absent in existing literature is also a comparative study between the two aggregation strategies, a gap that is being bridged for classification tasks \cite{gaudreault2024empirical,suhaimi2022comparative}. Our study of existing macro-averaging implementations (ClusterCrit) reveals that only uniform sampling is employed, often used for the application on large datasets. We observe, however, that uniform sampling cancels the benefits of macro-averaging (i.e., in cluster-imbalanced spaces), because the measure reduces effectively to micro-averaging. We address this gap by proposing an alternative sampling approach, well-suited to macro-averaging. 

\section{The Silhouette Coefficient} \label{sec:silhouette}

Data clustering is one of the most fundamental unsupervised learning tasks with numerous applications in computer science, among many other scientific fields~\cite{jain2010data,ezugwu2022comprehensive}. Although a strict definition of clustering may be difficult to establish, a more flexible interpretation can be stated as follows: \textit{Clustering is the process of partitioning a set of data points into groups (clusters), such that points of the same group share ``common'' characteristics while ``differing'' from points of other groups}. Data clustering can reveal the underlying data structure and hidden patterns in the data. At the same time, it is a task that poses several challenges due to the absence of labels~\cite{jain1999data}, including the evaluation of clustering solutions. 

Assessing the quality of a clustering solution ideally requires human expertise~\cite{von2012clustering}. However, finding human evaluators could be hard, expensive and time-consuming (or even impossible for very large datasets). 
An alternative approach is to use clustering evaluation measures, which can be either external (supervised) or internal (unsupervised)~\cite{rendon2011internal}. The former, as the name suggests, use external information (e.g., classification labels) as the ground truth cluster labels. Well known external evaluation measures are Normalised Mutual Information (NMI)~\cite{estevez2009normalized}, Adjusted Mutual Information (AMI)~\cite{JMLR:v11:vinh10a}, Adjusted Rand Index (ARI)~\cite{hubert1985comparing,chacon2022minimum}, etc. 
External information, however, is not typically available in real-world scenarios. In such cases we resort to internal evaluation measures, which are solely based on information intrinsic to the data. Although other internal evaluation measures have been proposed~\cite{calinski1974dendrite,davies1979cluster}, we focus on the most commonly-employed, and successful one \cite{arbelaitz2013extensive}, which is the silhouette coefficient~\cite{rousseeuw1987silhouettes}.

\begin{figure}[t]
\centering
\scalebox{1.2}{
\begin{tikzpicture}
\def\clusterOneX{0}
\def\clusterOneY{0}
\def\clusterTwoX{2}
\def\clusterTwoY{2}
\def\clusterThreeX{4}
\def\clusterThreeY{-1}
\def\radius{0.75cm}
\def\xiX{0.35}
\def\xiY{0.25}
\fill[green!40] (\clusterOneX,\clusterOneY) circle (\radius);
\draw (\clusterOneX,\clusterOneY) node[above, yshift=\radius] {$C_I$};
\fill[caribbeangreen!40] (\clusterTwoX,\clusterTwoY) circle (\radius);
\draw (\clusterTwoX,\clusterTwoY) node[above, yshift=\radius] {$C_J$};
\fill[red!40] (\clusterThreeX,\clusterThreeY) circle (\radius);
\draw (\clusterThreeX,\clusterThreeY) node[above, yshift=\radius] {$C_K$};
\foreach \x/\y in {\clusterOneX+\xiX/\clusterOneY+\xiY,\clusterOneX-0.5/\clusterOneY-0.25,\clusterOneX+0.15/\clusterOneY-0.2,
                   \clusterTwoX-0.35/\clusterTwoY-0.1, \clusterTwoX+0.5/\clusterTwoY-0.35, \clusterTwoX+0.15/\clusterTwoY+0.15,
                   \clusterThreeX+0.00/\clusterThreeY+0.55, \clusterThreeX-0.15/\clusterThreeY-0.35, \clusterThreeX+0.25/\clusterThreeY-0.1}
    \fill (\x,\y) circle (2pt);

\fill[black] (\xiX,\xiY) circle (2pt) node[above] {$x_i$};
\foreach \x/\y in {\clusterOneX+\xiX/\clusterOneY+\xiY, \clusterOneX-0.5/\clusterOneY-0.25, \clusterOneX+0.15/\clusterOneY-0.2,
                   \clusterTwoX-0.35/\clusterTwoY-0.1, \clusterTwoX+0.5/\clusterTwoY-0.35, \clusterTwoX+0.15/\clusterTwoY+0.15,
                   \clusterThreeX+0.00/\clusterThreeY+0.55, \clusterThreeX-0.15/\clusterThreeY-0.35, \clusterThreeX+0.25/\clusterThreeY-0.1}
    \draw[black, dashed] (\xiX,\xiY) -- (\x,\y);
\end{tikzpicture}}
\caption{Illustration of the elements involved in the computation of the silhouette score $s(x_i)$ for a given data point $x_i$ that belongs to cluster $C_I$.}
\label{fig:Silhouette}
\end{figure}
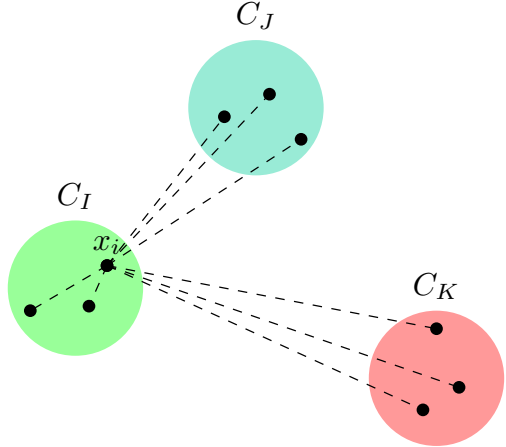
The silhouette coefficient~\cite{rousseeuw1987silhouettes} is a measure to assess clustering quality, which does not depend on external knowledge and that does not require ground truth labels. A good clustering solution, according to this measure, assumes compact and well-separated clusters. Formally, given a dataset $X = \{x_1, ..., x_N\}$ that is partitioned by a clustering solution $f: X \rightarrow \{C_1, ..., C_K\}$ into $K$ clusters, the silhouette coefficient for point $x_i \in X$ is based on two values, the inner and the outer cluster distance.
The former, denoted as $a(x_i)$, is the average distance between $x_i$ and all other points within the cluster $C_I$ that $x_i$ belongs to (i.e., $f(x_i) = C_I$):
\begin{equation}
\label{eq:ai}
a(x_i) = \frac{1}{|C_I| - 1} \sum_{x_j \in C_I, i \neq j} d(x_i, x_j),
\end{equation}
where $|C_I|$ represents the cardinality of cluster $C_I$ and $d(x_i, x_j)$ is the distance between $x_i$ and $x_j$. The $a(x_i)$ value quantifies how well the point $x_i$ fits within its cluster. For example, in Figure~\ref{fig:Silhouette}, $a(x_i)$ measures the average distance of $x_i$ to the points in its cluster $C_I$. A low value of $a(x_i)$ indicates that $x_i$ is close to the other members of that cluster, suggesting that $x_i$ is probably grouped correctly. Conversely, a higher value of $a(x_i)$ indicates that $x_i$ is not well-placed in that cluster. In addition, the silhouette score requires the calculation of the minimum average outer-cluster distance $b(x_i)$ per point $x_i$, defined as:
\begin{equation}
\label{eq:bi}
b(x_i) = \min_{C_J \neq f(x_i)}\frac{1}{|C_J|} \sum_{x_j \in C_J} d(x_i, x_j).
\end{equation}
A large $b(x_i)$ value indicates that $x_i$ significantly differs from the points of the closest cluster. In Figure~\ref{fig:Silhouette}, the closest cluster (which minimises $b(x_i)$) is $C_J$. Considering both $a(x_i)$ and $b(x_i)$, the silhouette score of $x_i$ is defined as:
\begin{equation}
\label{eq:si}
s(x_i) = \frac{b(x_i) - a(x_i)}{\max\left\{a(x_i), b(x_i)\right\}}.
\end{equation}
It is evident that the silhouette score $s(x_i)$, defined in Eq.~\ref{eq:si} for a data point $x_i$, falls within the range $ -1\leq s(x_i)\leq 1$. A value close to 1 indicates that the data point $x_i$ belongs to a compact, well-separated group. In contrast, a value close to -1 suggests that another cluster assignment for that data point would have been a better option.  

\section{Methods}\label{sec:methods}
The Silhouette Coefficient provides a score that grades the cluster assignment of a data point. To obtain a single score for all the points $x \in X$, the typical approach is micro-averaging (see \S\ref{sec:literature}) that averages all the individual scores. The alternative macro-averaging approach averages first the scores per cluster, and then (macro) average the latter.

\subsection{Micro-averaged Silhouette: \emph{The typical index}} 
Micro-averaging silhouette at the point level, or sample mean in short, is defined as follows: 
\begin{equation}
\label{eq:Silhouette}
S_{micro}(X) = \frac{1}{N} \sum_{x_i \in X} s(x_i).
\end{equation}

This is the originally proposed averaging strategy by the study that introduced silhouette~\cite{rousseeuw1987silhouettes}, the one adopted by \href{https://scikit-learn.org/stable/modules/generated/sklearn.metrics.silhouette_score.html}{\textsc{scikit-learn}}, and it is the typical approach employed in literature \cite{batool2021clustering}. However, we show in \S\ref{sec:experiments} that it is not effective in the case of imbalanced clusters, which is a very common property of real-world datasets. 

\subsection{Macro-averaged Silhouette: \emph{The overlooked index}}
When clusters are perfectly balanced, the sample mean is a reasonable aggregation strategy. The assumption of perfectly balanced clusters, however, cannot be guaranteed in the real world, where clusters are often imbalanced. In such cases, and when small clusters also matter (e.g., diagnostic reports about rare medical diseases), we argue that micro-averaging is not effective while macro-averaging is robust. This issue is known in fields such as supervised learning \cite{gaudreault2024empirical}, but it has not been studied yet for clustering. 

To compute the macro-Silhouette, a score $S_c$ is computed for each cluster $C_i$ as follows:
\begin{equation}
\label{eq:cluster-sil}
S_{C}(C_i) = \frac{1}{|C_i|}\sum_{x_i \in C_i} s(x_i).
\end{equation}
This score measures how compact and well separated a cluster is given a clustering solution. For $K$ clusters in that solution, we end up with a set of $K$ cluster silhouette values \{$S_{C_1}, \ldots, S_{C_K}$\}. The average of these $K$ scores, defined as the macro-averaged Silhouette, can be used to assess the dataset clustering and is more formally defined as follows:
\begin{equation}
\label{eq:Macro-Average-Silhouette}
S_{macro}(X) = \frac{1}{K} \sum_{k=1}^{K} S_C(C_i).
\end{equation}

We note that macro-averaging assumes equal weight between the clusters, but other approaches also apply. The weighted average, for example, where weights reflect the support (i.e., the number of points per cluster, normalised), is closer to micro-averaging in nature. Furthermore, other statistics could be applied, such as the max (or the min), capturing the most (least) compact and well (bad) separated cluster. 

\subsection{Per-cluster sampling: \emph{Efficient and robust macro-Silhouette}}\label{ssec:sampling}
The computation of the silhouette coefficient for all the $N$ points in a dataset requires the computation of a pairwise distance matrix at the cost of $\mathcal{O}(N^{2})$ operations. This is demanding in terms of computational and space complexity and, hence, not scalable for large datasets~\cite{capo2023fast}. The typical approach to tackle this problem is to compute the silhouette score using a uniformly selected subsample of the dataset. 

In a cluster-imbalanced problem, the typical (uniform across data points) sampling may favour the major cluster and may even disregard completely one of the minor clusters. We argue that this practice contradicts the nature of macro-averaging, which assumes that clusters are equally weighted when averaged. To solve this problem, when macro-averaging is aimed, we propose that sampling takes place per cluster, following the macro-averaging spirit. 

More specific, we create a subsample of size $L$ for computing the macro-averaged silhouette score by uniformly selecting a subset of $L/K$ points from each cluster $C_i$, where $i=1,\ldots,K$. In this way, we ensure that all the clusters contribute a sufficient number of data points to the subsample, preserving the robustness of macro-Silhouette.

\section{Experiments}\label{sec:experiments}
\begin{figure}
    \centering
    \includegraphics[width=.9\textwidth]{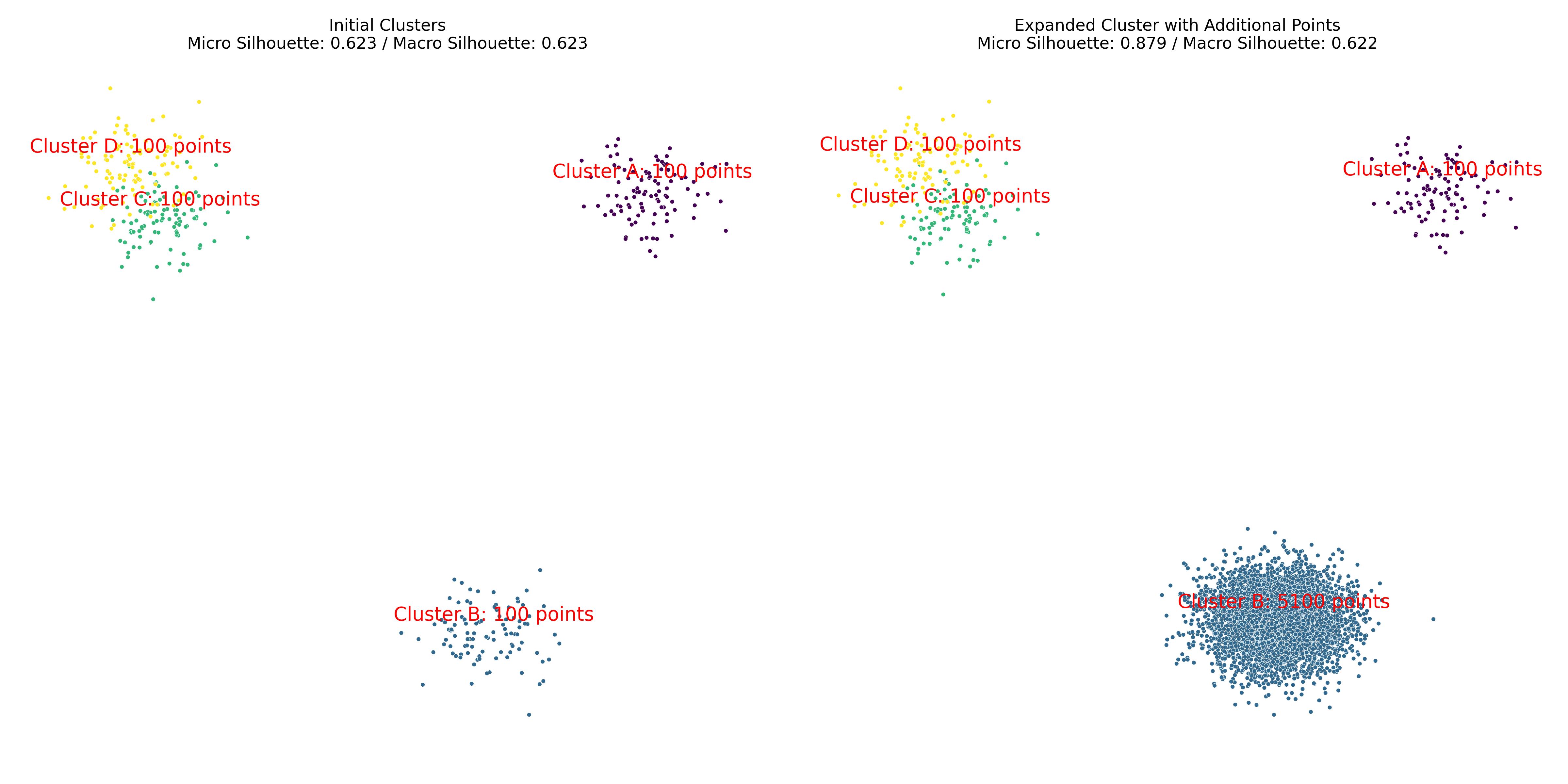}
    \caption{Synthetic dataset, shown on the left, with four equibalanced clusters. The same space is shown on the right, but the relatively distant cluster B now comprises 5,000 points more, yielding a heavy cluster-imbalance. Silhouette is reported per dataset per aggregation strategy. Micro-averaging increases unreasonably by a large margin.}
    \label{fig:synthetic}
\end{figure}

\subsection{Analysis on synthetic data}\label{sec:synthetic}
We created a synthetic dataset consisting of four Gaussian clusters, each with 100 points and a variance of 0.1, shown on the left of Figure~\ref{fig:synthetic}. The micro- and macro-averaged Silhouette scores are both 0.623, a score that is far from perfect due to the overlapping clusters C and D. The same space is shown on the right, but the lower distant cluster B is now populated with 5,000 points more, resulting in a heavy cluster imbalance. Specifically, the ratio of the smallest to the largest cluster for this dataset is $r=0.02$. 

\begin{figure}
    \centering
    \includegraphics[width=.7\textwidth]{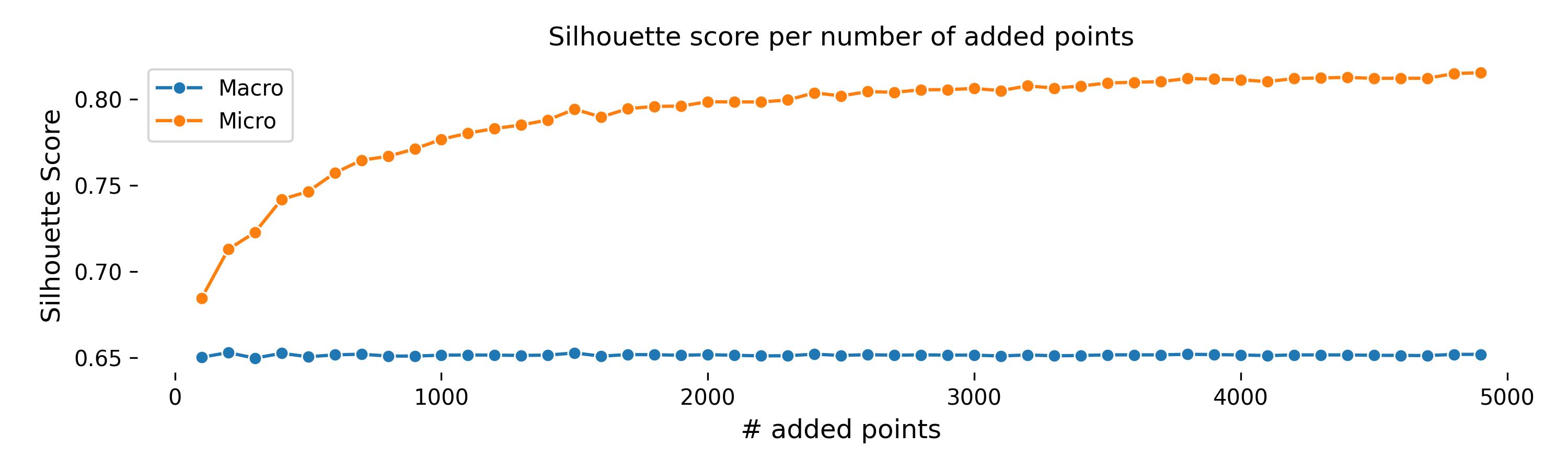}
    \caption{Silhouette score of the dataset of Figure~\ref{fig:synthetic}, micro- and macro-averaged, for a varying number of points added.}
    \label{fig:synthetic_grow}
\end{figure}

When moving from the (balanced) space on the left to the (imbalanced) space on the right, micro changes considerably. That is because the points added to the distant cluster B directly influence the point-level average value upwards (i.e., a relative increase of 41\%). The overlap of the minor clusters C and D is less important in this case, when compared to the well-separated major cluster B. By contrast, macro-average remains robust to this change, because each cluster is equally weighted in the average, disregarding their size. As is shown in Figure~\ref{fig:synthetic_grow}, the sensitivity of micro-averaging to imbalance becomes apparent early on and it could continue increasing if we continued adding points.   

\begin{figure}[h]
    \centering
    \includegraphics[width=.7\textwidth]{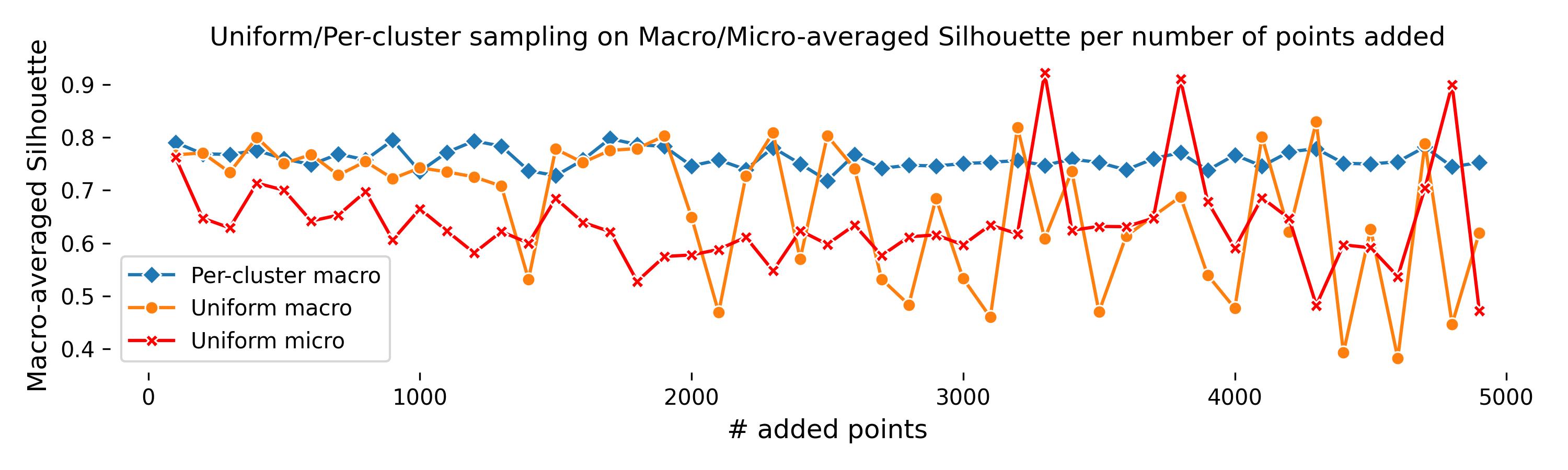}
    \caption{Macro-averaged Silhouette score, computed on uniform and per-cluster samples of 100 points, as the size of cluster B of Figure~\ref{fig:synthetic} increases.}
    \label{fig:sampling}
\end{figure}

\paragraph{Per-cluster sampling}
Using the same dataset, we assess the robustness of the proposed per-cluster sampling. This is clearly shown in Figure~\ref{fig:sampling}, by comparing the typical uniform and the proposed per-cluster sampling of the macro-averaged Silhouette score. The more the imbalance, as we move to the right of the Figure, the more the fluctuations of the macro-averaged Silhouette score computed on a uniform sample of 100 points. The per-cluster sampling, on the other hand, remains robust. Similar fluctuations are observed on uniform sampling and micro-averaging (in red).

\begin{figure}
    \centering
    \includegraphics[width=.9\textwidth]{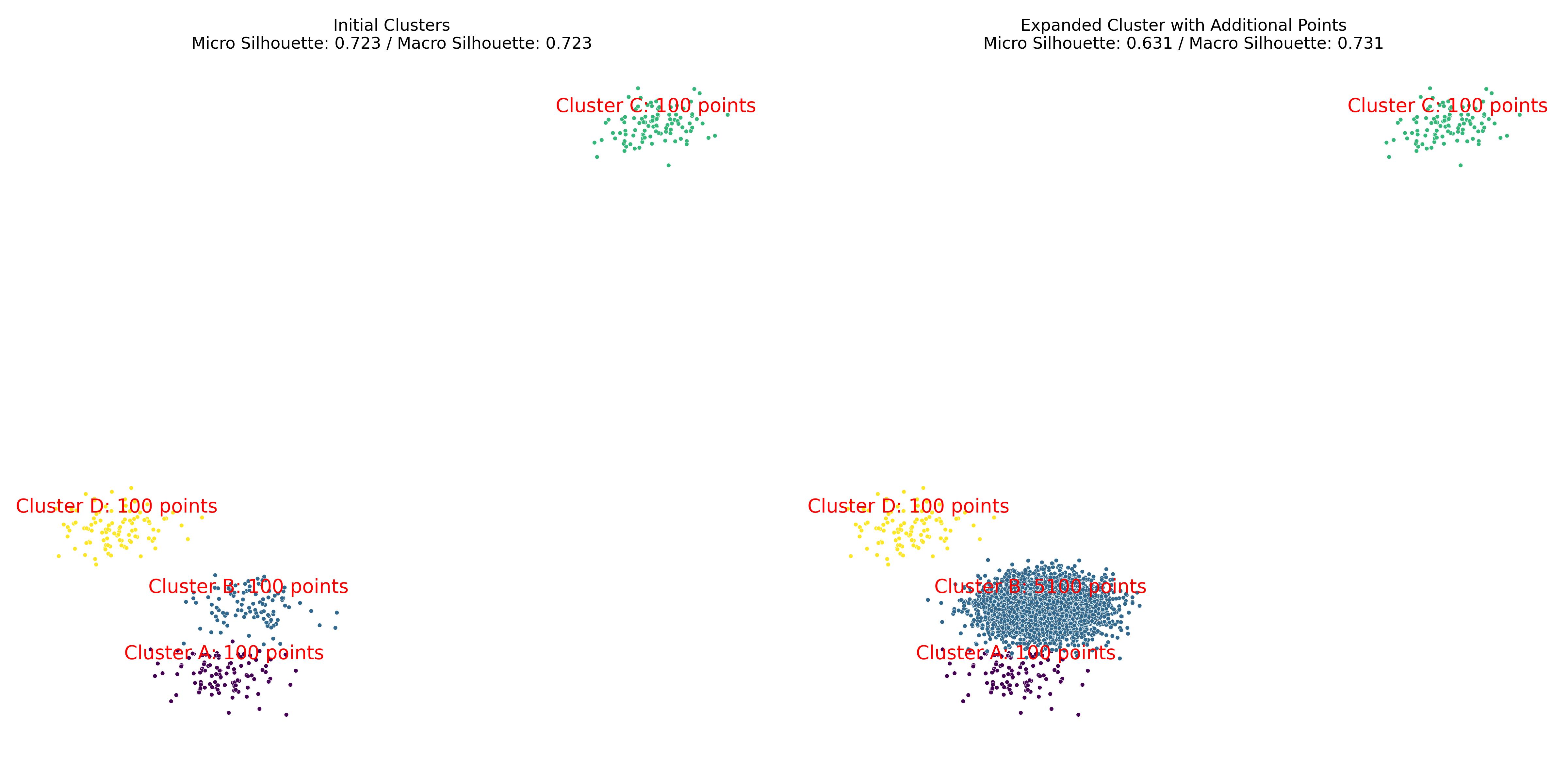}
    \caption{Synthetic dataset, equibalanced as in Figure~\ref{fig:synthetic} on the left, but now cluster B (to which we add 5,000 points on the right) is very close to clusters D and A.}
    \label{fig:synthetic2}
\end{figure}

\paragraph{Major overlapping cluster} When adding points to a distant cluster, we observe an increase of micro-averaged Silhouette as opposed to its macro-averaged counterpart. The situation is different, however, when the cluster we add points to is close to others. Figure~\ref{fig:synthetic2} depicts this space, where we observe that micro-averaged Silhouette drops in value in this case. Macro-Silhouette, on the other hand, again remains robust. This is mainly because we add points that overlap with nearby clusters, reducing the overall score instead of increasing it, as was the case when we added points to a distant cluster (Figure~\ref{fig:synthetic_grow}).

\begin{figure*}[h!]
\centering
\begin{subfigure}{0.45\linewidth}
\centering
\includegraphics[width=\linewidth]{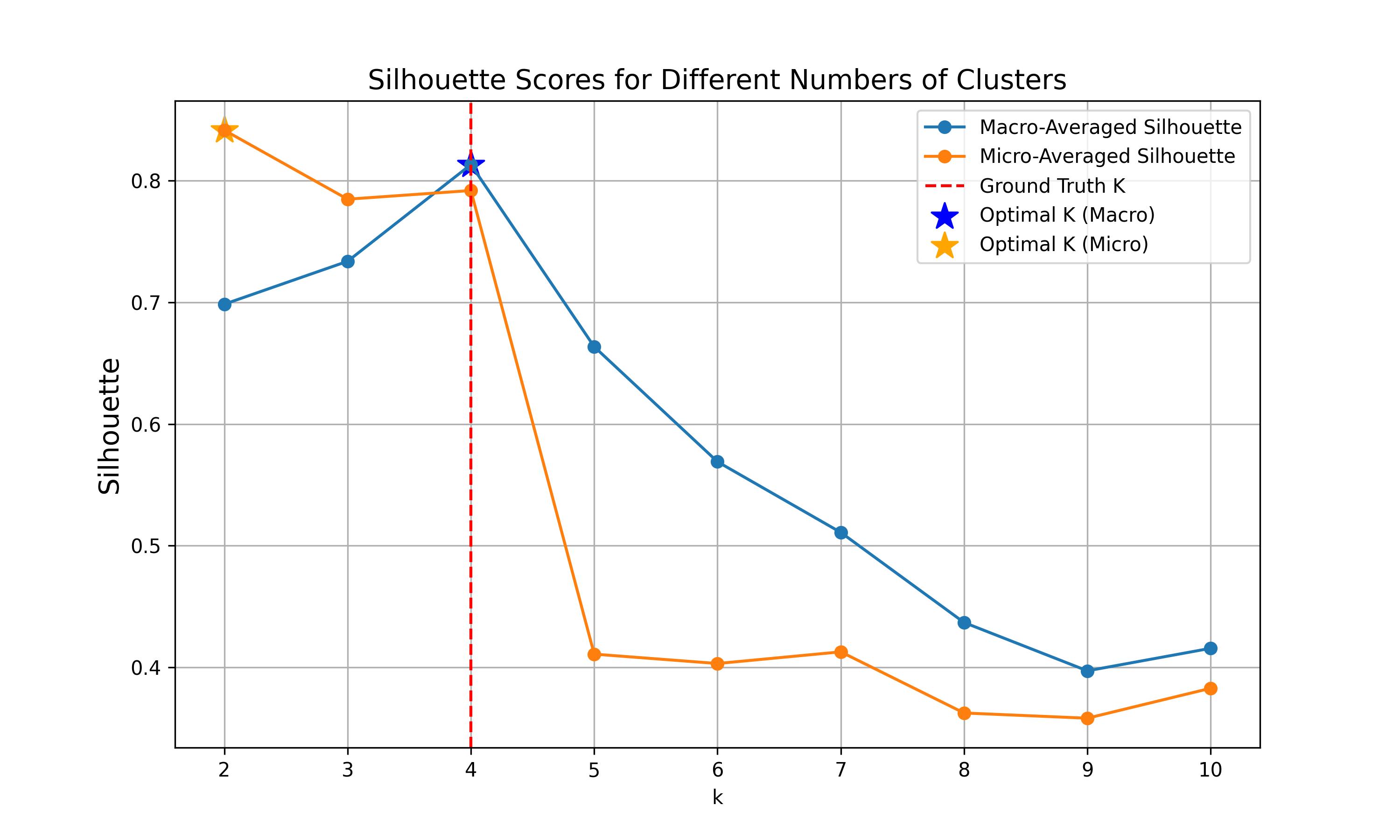}
\caption{Imbalanced}
\end{subfigure}
\begin{subfigure}{0.45\linewidth}
\centering
\includegraphics[width=\linewidth]{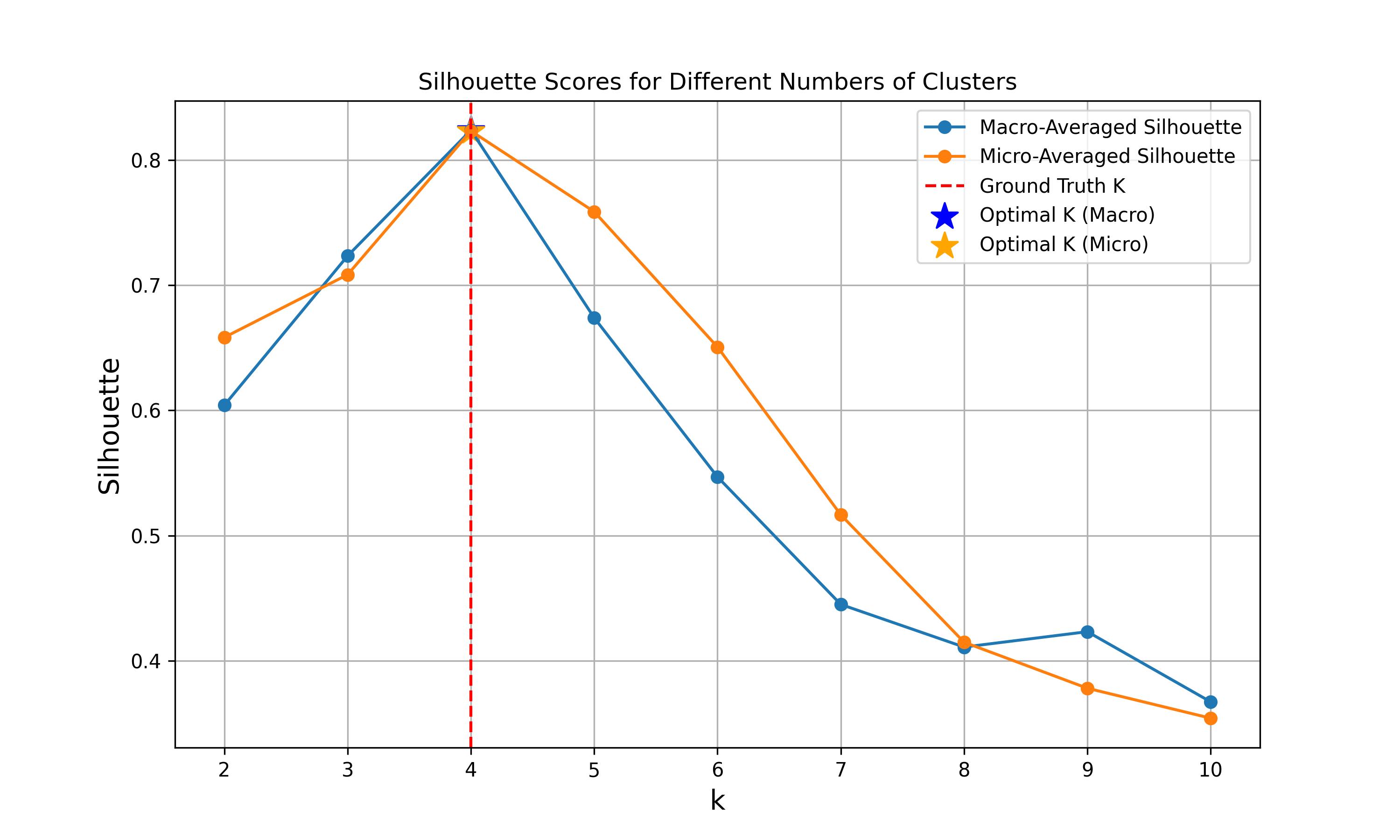}
\caption{Balanced}
\end{subfigure}
\caption{Estimating the optimal number of clusters (shown with star) when using micro (orange) and macro (blue) averaged Silhouette. We use both, an imbalanced dataset of four isotropic Gaussians, and an undersampled (balanced) version of the same space. A red vertical line shows the ground truth number of clusters. }
\label{fig:find_k_synth}
\end{figure*}

\paragraph{Estimating $k$ (the Silhouette method)} 
Silhouette has been suggested as an alternative to the problematic ``elbow'' method when estimating the number of clusters \cite{schubert2023stop}. That is, the quality of a clustering solution (e.g., the predictions of KMeans) is evaluated for different numbers of clusters. The number that leads to the highest Silhouette score is chosen as the optimal one. We applied this method on a synthetic imbalanced dataset of four isotropic Gaussian blobs,\footnote{\url{https://scikit-learn.org/stable/modules/generated/sklearn.datasets.make_blobs.html}} three with 100 points and one with 2,300. By undersampling from the major cluster, we also produce an equibalanced version of this space where clusters comprised 100 points each. By applying KMeans, then, with $k$ ranging from 2 to 10, we measured the micro and macro averaged Silhouette per space per $k$, using uniform (for micro) and per-cluster (for macro) sampling of 100 points. As is shown in Figure~\ref{fig:find_k_synth}, macro-averaged Silhouette reaches a maximum (blue star) on the ground truth number of clusters (red line) in the imbalanced dataset. Micro-averaged Silhouette is maximised for a different number of clusters. In the undersampled (balanced) version of this dataset, shown on the right, both strategies reach their maximum on the ground truth number of clusters, i.e., four, where the red vertical line is. This result can be explained by the lack of robustness of uniform-sampled micro-averaged Silhouette (see Figure~\ref{fig:sampling}), which is deceiving when estimating the number of clusters. The robust per-cluster-sampled macro-averaged Silhouette, on the other hand, is not affected.

\subsection{Application on real-world datasets}\label{ssec:real}
We employed eight real-world datasets \cite{Dua:2019} of various types (numeric, time-series, images), sizes (from 150 to more than 500,000 items), dimensionality, and with a varying cluster imbalance. To estimate the latter, we computed the ratio ($r$) of the size of the smallest to the largest cluster. Table~\ref{tab:datasets} displays these datasets, sorted according to their imbalance, which ranged from high ($r=0.03$) to low ($r=1.00$). 

These eight datasets are summarised below:
\begin{itemize}
    \item \pendigits comprises 10,992 pen-based handwritten digits (from $0$ to $9$). Each data item is represented by a $16$-dimensional vector containing pixel coordinates. \digits also comprises (1,797) images of handwritten digits, but each item is an image of 8x8 pixels, resulting in $d=64$ features. 
    \item \cover contains cartographic variables for predicting forest cover types. The dataset includes 581,012 samples and 54 features, such as elevation, aspect, slope, and soil type. The cover types are classified into seven categories and it is a highly imbalanced dataset ($r=0.03$).
    \item \gas consists of measurements from 16 chemical sensors exposed to different gases over a period of several months. The dataset includes 13,910 samples and 128 features. It is used for studying the drift in sensor responses over time and developing algorithms for sensor calibration.
    \item \wine contains the results of a chemical analysis of wines grown in the same region in Italy but derived from three different cultivars. The analysis determined the quantities of 13 constituents found in each of the three types of wines. 
    \item \iris comprises the lengths and widths of the sepals and petals of iris flowers. 
    \item \glass contains the chemical compositions of glass samples, which are classified into seven types of glass. 
    \item \tcga is a collection of gene expression profiles obtained from RNA sequencing of various cancer samples. It includes $801$ data instances, clinical information, normalised counts, gene annotations, and $6$ cancer types' pathways. 
    \item \mice consists of the expression levels of 77 proteins/protein modifications that produced detectable signals in the nuclear fraction of the cortex. It includes $1,080$ data points and $8$ eight classes of mice based on the genotype, behaviour, and treatment characteristics.
\end{itemize}

\paragraph{Evaluation metrics}
We compute micro and macro Silhouette scores, as internal validation measures. We also employ external validation measures that use the ground truth labels to assess the clustering solution, presented for completeness. The normalized mutual information (NMI) score measures the similarity between two clusterings by normalizing the mutual information score \cite{vinh2009information}. A higher score indicates a better match between the cluster labels and the ground truth labels. The adjusted mutual information (AMI) adjusts MI for chance groupings \cite{vinh2009information}. It measures the agreement between two clustering assignments and is normalized against the entropy of the labels to yield a score between 0 and 1. The adjusted rand index (ARI) measures the similarity between two data clusterings by adjusting Rand Index (RI) to account for the chance grouping of elements \cite{hubert1985comparing}. The score ranges from -1 to 1, with higher values indicating better clustering performance. 

\begin{table}[ht]
    \centering
    \caption{Real-world datasets of varying dimensionality ($d$), size ($N$), number of clusters ($k$), imbalance ratio of smallest to largest cluster ($r$). The average macro (MaS) and micro (MaS) Silhouette score, sampled uniformly and per-cluster respectively, is reported across three runs (st. error of mean), along with NMI, ARI, AMI. Sorted by $r$.}

\begin{tabular}{llllcccccccc}
\toprule
     Dataset &        Type &      N &   d &  k &    r &       MaS &       MiS &  NMI &  ARI &  AMI \\
\midrule
        Iris &     Numeric &    150 &   4 &  3 & 1.00 & 0.46±0.01 & 0.46±0.01 & 0.66 & 0.62 & 0.66 \\
      Digits &       Image &   1797 &  64 & 10 & 0.95 & 0.11±0.01 & 0.13±0.01 & 0.69 & 0.56 & 0.69 \\
   Pendigits & Time-series &  10992 &  16 & 10 & 0.92 & 0.24±0.01 & 0.23±0.01 & 0.67 & 0.53 & 0.67 \\
Mice Protein &     Numeric &   1080 &  77 &  8 & 0.70 & 0.13±0.01 & 0.12±0.01 & 0.26 & 0.14 & 0.25 \\
        Wine &     Numeric &    178 &  13 &  3 & 0.68 & 0.30±0.01 & 0.29±0.01 & 0.86 & 0.88 & 0.86 \\
  Gas Sensor & Time-series &  13910 & 128 &  6 & 0.55 & 0.22±0.03 & 0.27±0.01 & 0.19 & 0.07 & 0.19 \\
       Glass &     Numeric &    214 &   9 &  6 & 0.12 & 0.20±0.01 & 0.31±0.01 & 0.32 & 0.17 & 0.29 \\
   Covertype &     Numeric & 110393 &  54 &  7 & 0.03 & 0.25±0.01 & 0.08±0.01 & 0.13 & 0.05 & 0.13 \\
\bottomrule
\end{tabular}

    \label{tab:datasets}
\end{table}

\paragraph{Experimental settings}
Missing values in datasets were replaced with the mean value of the respective feature.\footnote{\url{https://scikit-learn.org/stable/modules/generated/sklearn.impute.SimpleImputer.html}.}
We standardized all the features per dataset, by removing the mean and scaling to unit variance,\footnote{\url{https://scikit-learn.org/stable/modules/generated/sklearn.preprocessing.StandardScaler.html}.} 
to avoid numerical instabilities in the computations~\cite{celebi2013comparative}. 
We trained KMeans per dataset, using the ground truth number of clusters and selecting initial cluster centroids with KMeans++.

\subsubsection{Results with fixed ground truth number of clusters}
Table~\ref{tab:datasets} presents the evaluation results of KMeans across datasets, setting $k$ to the ground truth number of clusters of each dataset. The external validation scores (NMI, ARI, AMI) are high for balanced datasets ($r>0.9$), they drop in two out of three mild-imbalanced ones ($0.5\leq r \leq0.7$), and they are overall low for both highly imbalanced datasets. 
When computing Silhouette, we employ uniform (for micro) and per-cluster (for macro) sampling of 100 points per dataset, repeating three times and reporting the average and the standard error of the mean. 
We observe that the two indices are very close to each other in the balanced and mild imbalanced zones, with any differences not exceeding the standard error of the mean. By contrast, in the highly imbalanced zone, the two indices are far from each other. The sensitivity of micro-averaging to cluster imbalance (\S\ref{sec:synthetic}) can explain the observed difference between the two indices. When MiS is greater (e.g., \glass), a major distant cluster may be present while when MaS is greater than MiS (e.g., \cover), a major cluster may be close to other ones.

\subsubsection{Results on $k$-estimation}
\begin{figure*}[h!]
\centering
\begin{subfigure}{0.45\linewidth}
\centering
\includegraphics[width=\linewidth]{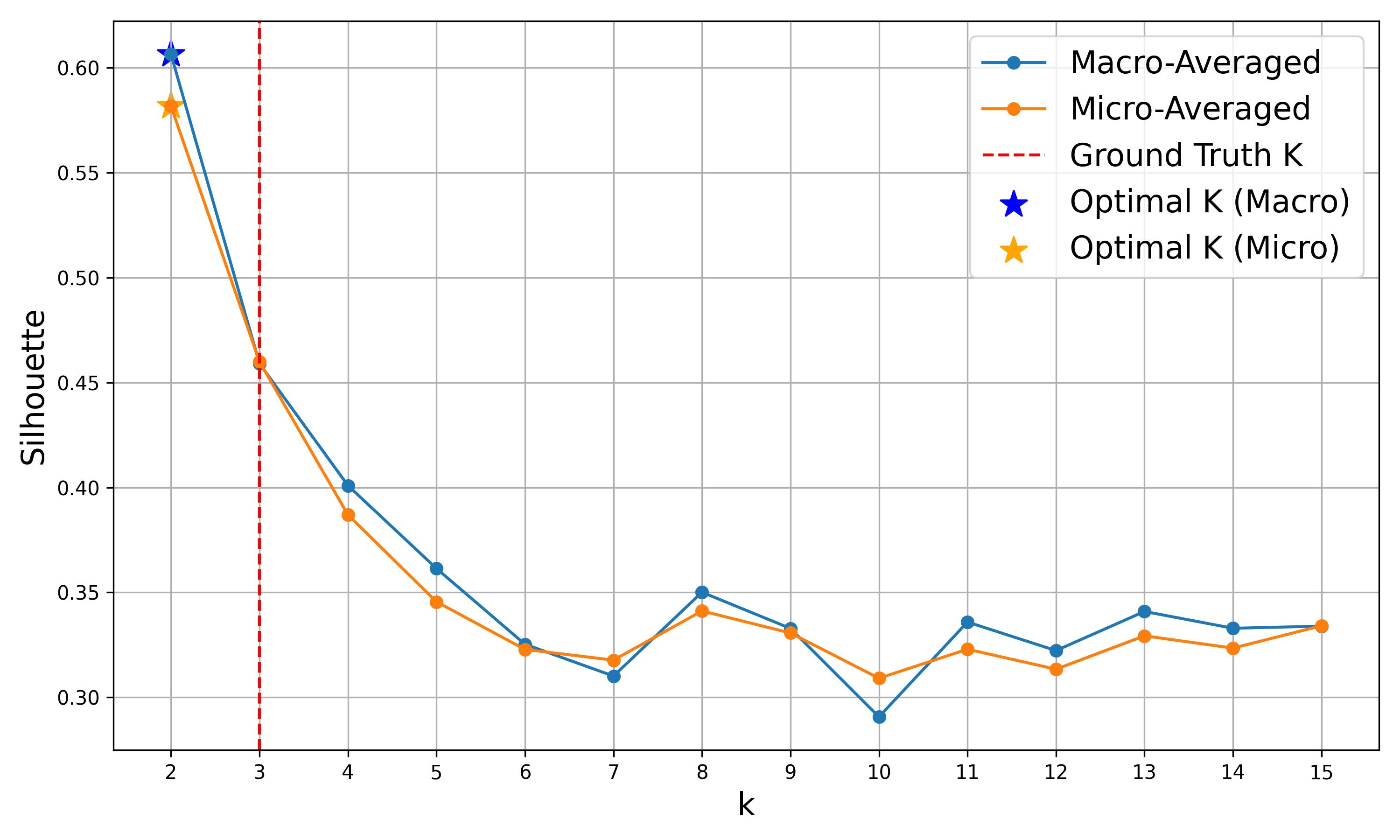}
\caption{\iris}
\end{subfigure}
\begin{subfigure}{0.45\linewidth}
\centering
\includegraphics[width=\linewidth]{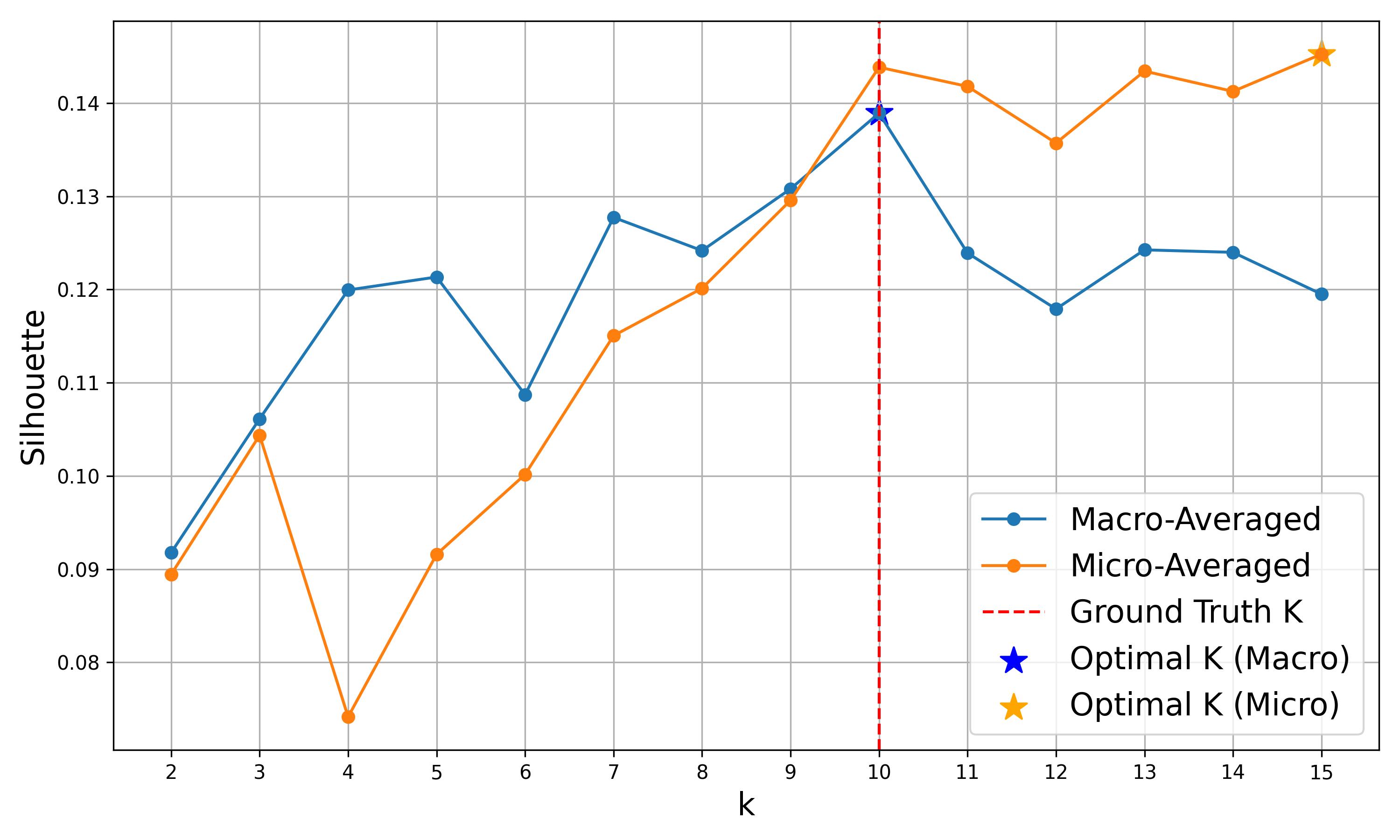}
\caption{\digits}
\end{subfigure}
\begin{subfigure}{0.45\linewidth}
\centering
\includegraphics[width=\linewidth]{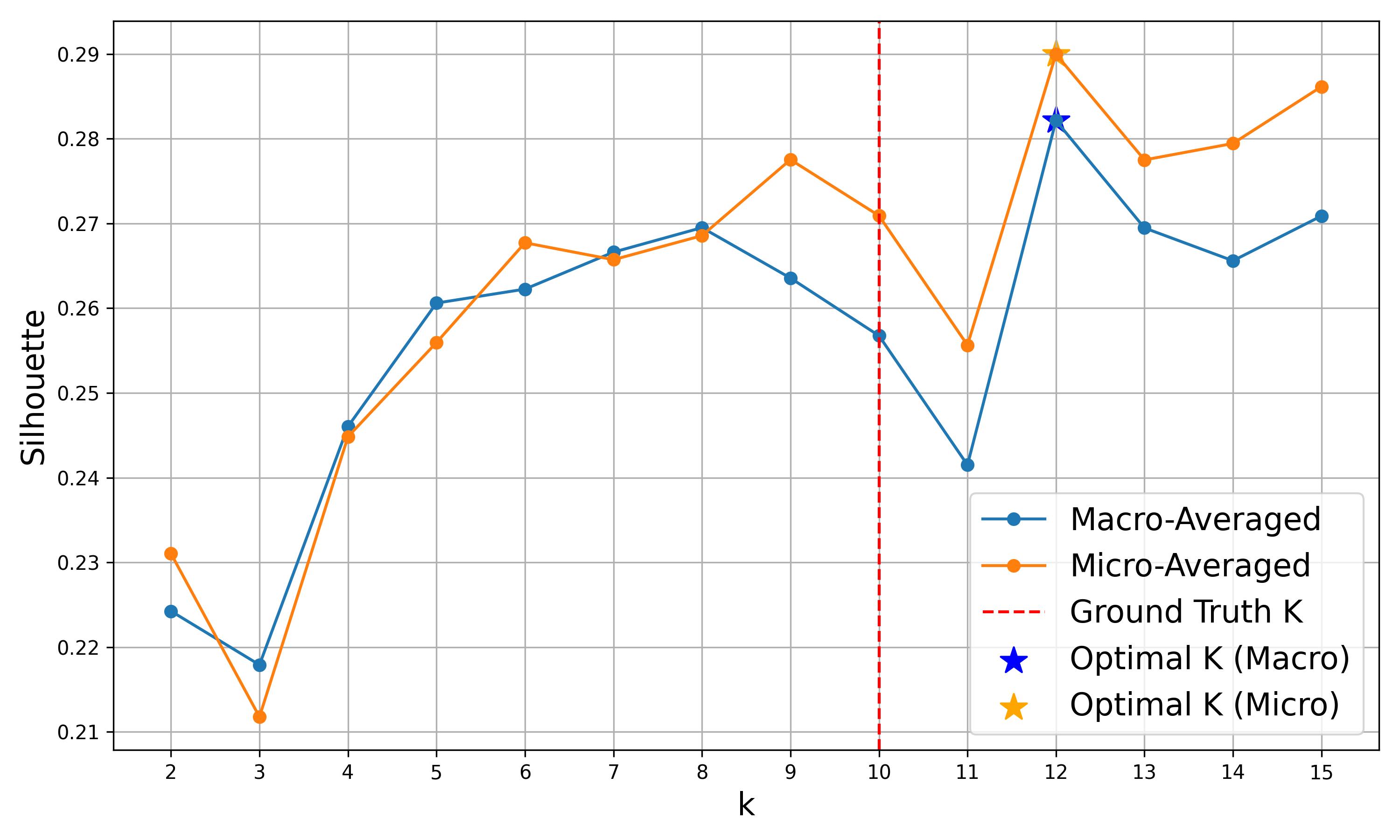}
\caption{\pendigits}
\end{subfigure}
\begin{subfigure}{0.45\linewidth}
\centering
\includegraphics[width=\linewidth]{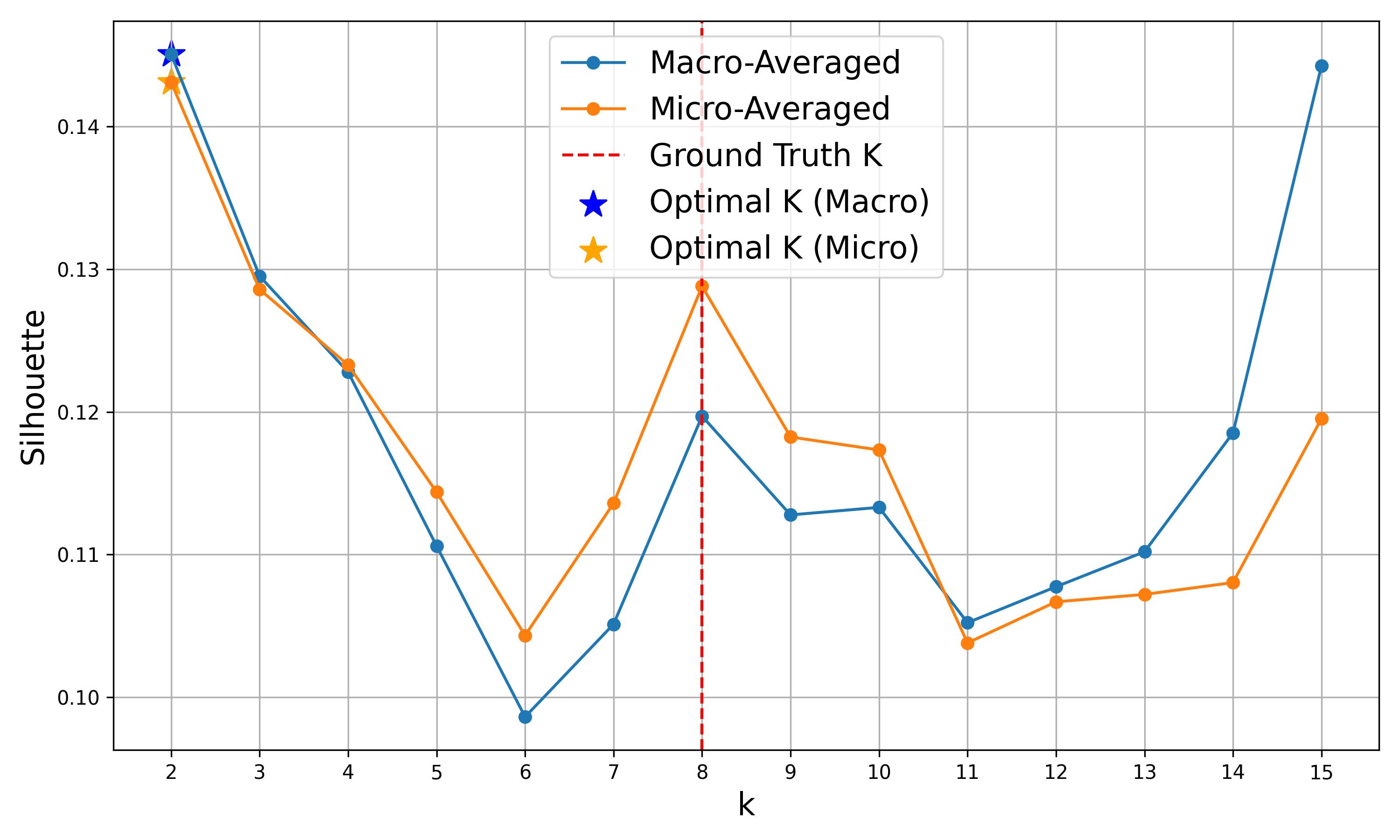}
\caption{\mice}
\end{subfigure}
\begin{subfigure}{0.45\linewidth}
\centering
\includegraphics[width=\linewidth]{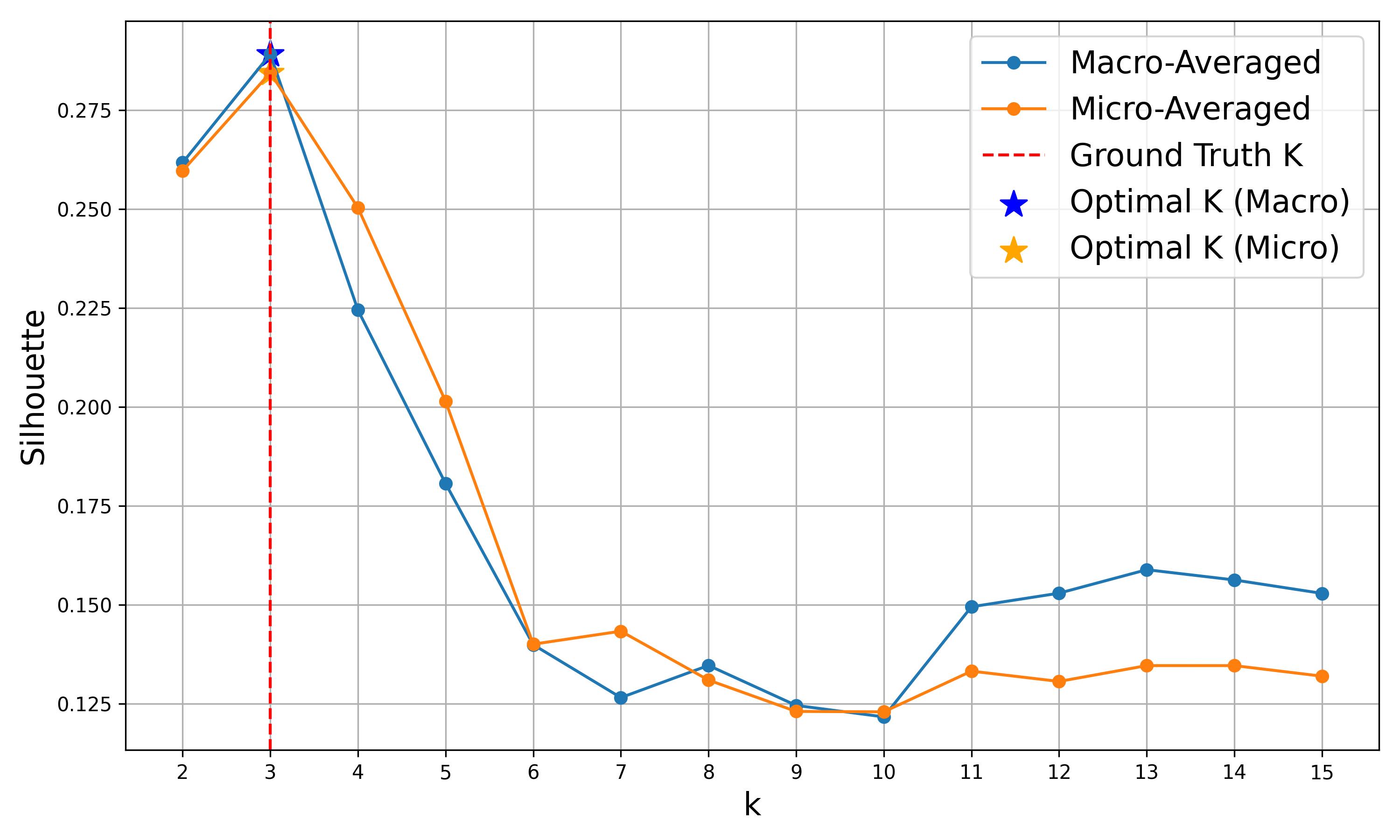}
\caption{\wine}
\end{subfigure}
\begin{subfigure}{0.45\linewidth}
\centering
\includegraphics[width=\linewidth]{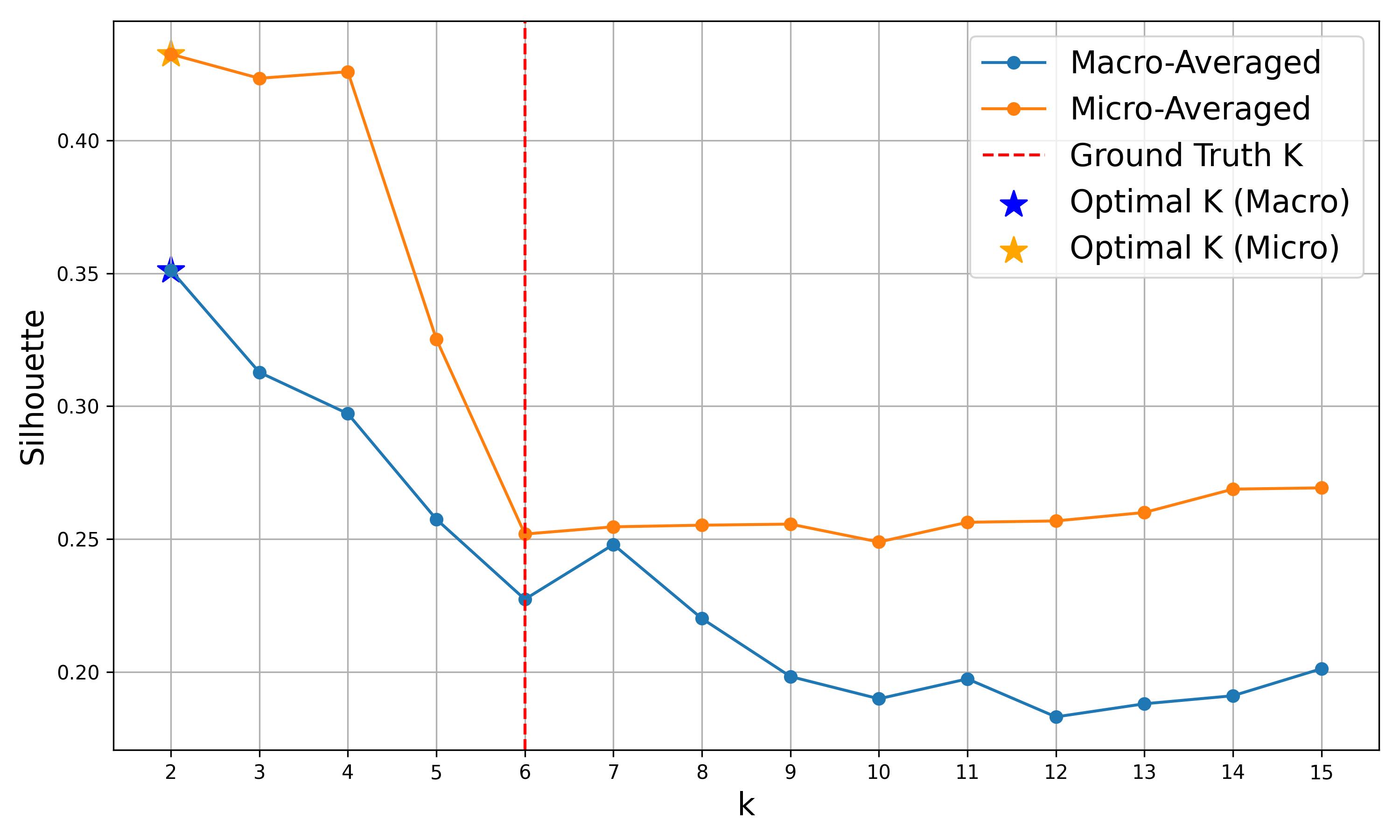}
\caption{\gas}
\end{subfigure}
\begin{subfigure}{0.45\linewidth}
\centering
\includegraphics[width=\linewidth]{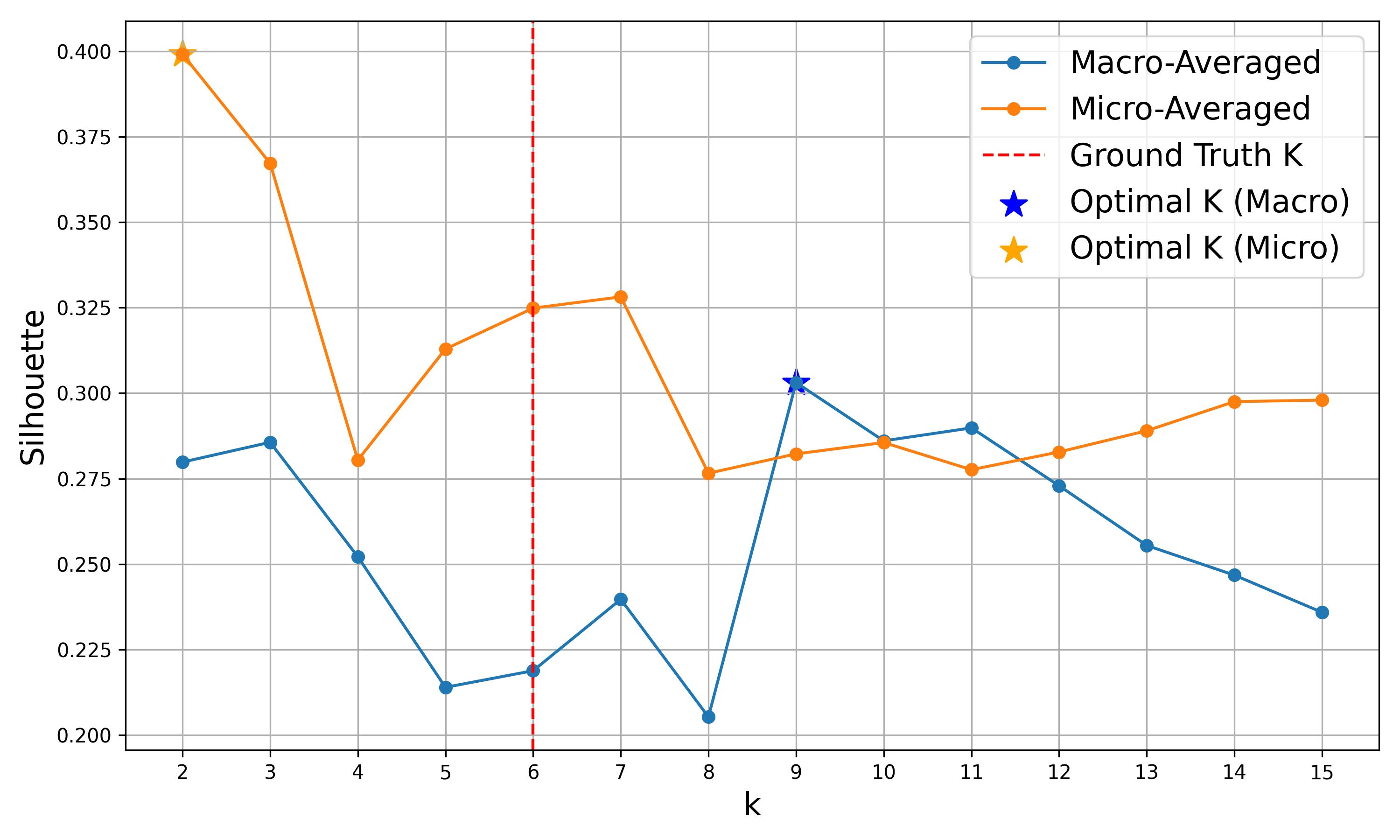}
\caption{\glass}
\end{subfigure}
\begin{subfigure}{0.45\linewidth}
\centering
\includegraphics[width=\linewidth]{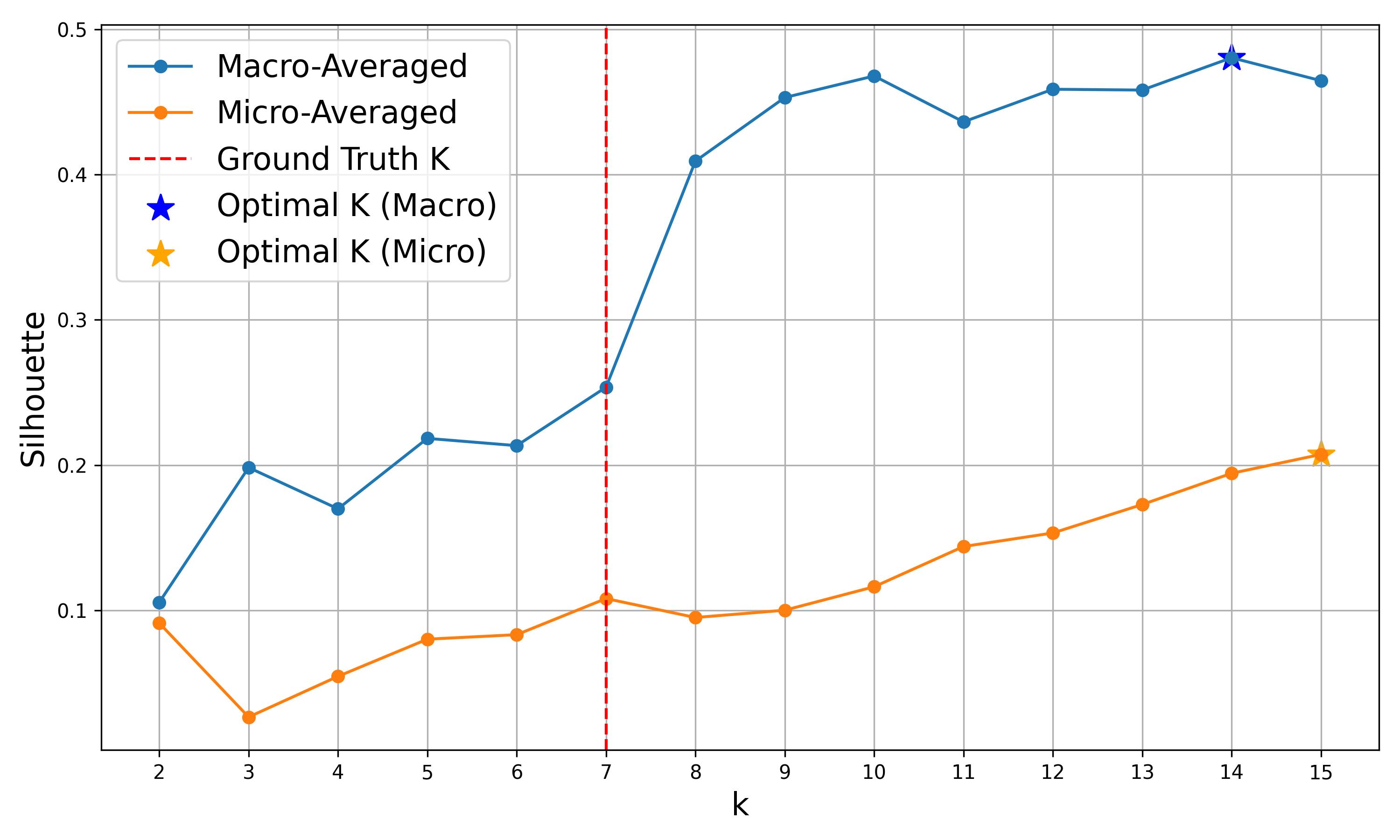}
\caption{\cover}
\end{subfigure}
\caption{Micro (blue) and macro (blue) averaged Silhouette per dataset on clusterings produced with KMeans for varying $k$ values as a means to select the optimal number of clusters (shown with a star). The ground truth number of clusters is shown with a vertical red line.}
\label{fig:optk}
\end{figure*}

We experimented also with estimating $k$ using the Silhouette method. That is by applying clustering for various $k$ values and assessing the two Silhouette aggregation strategies regarding their ability to yield a maximum score for the ground truth number of clusters. The ground truth number of clusters is shown with $k$ in Table~\ref{tab:datasets} and depicted with a red vertical line in Figure~\ref{fig:optk}. In Figure~\ref{fig:optk}, we see that micro and macro averaging yield the same optimal $k$ in five out of eight datasets, all balanced (\iris, \pendigits) or mildly imbalanced (\mice, \wine, \gas). In the heavily imbalanced datasets of \glass and \cover, as expected, the optimal $k$ differs. Overall, the macro-average yields an optimal $k$ that is the same as the ground truth $k$ in two datasets (\digits and \wine) while typical micro-averaging yields an optimal $k$ that is the same as the grand truth in one (\wine).


\section{Discussion}

Our study revisited the aggregation strategy for Silhouette Coefficient, which is widely-used as an internal evaluation measure for clustering solutions. Based on our findings, we argue that the aggregation strategy should not be selected arbitrarily nor should one trust the typical approach, which is micro-averaging. By contrast, we suggest that the choice between micro and macro-averaging Silhouette should be based on the nature of the dataset and the domain at hand. 

\paragraph{Silhouette aggregation and dataset} As was shown in \S\ref{sec:experiments}, the typical micro-averaged Silhouette score is vulnerable to cluster imbalance while the rarely-used macro-averaged Silhouette is far more robust. This means that when there is indication of an imbalanced dataset, it is macro-averaged Silhouette that should be trusted for evaluating clustering solutions. This is the case for the two most imbalanced real-world datasets used in this work. \glass achieves a lower macro-average compared to micro, which could be explained by a major distant cluster, as in Figure~\ref{fig:synthetic}. \cover, on the other hand, achieves a higher macro-average, compared to micro, which fits the synthetic example of Figure~\ref{fig:synthetic2} with a major cluster being nearby others. 

\paragraph{Silhouette aggregation per domain} The choice of the aggregation strategy could depend also on the domain at hand. In predictive maintenance, for example, major faults are of much higher importance compared to rare events \cite{pavlopoulos2024automotive}. That is because an accurate estimation of the frequency of the problem may allow better logistics and administration for the company (e.g., early orders, select appropriate workstations, etc.). In this case, micro-Silhouette should be the selected index, if clustering was applied. On the other hand, biomedical clustering applications would likely select macro-Silhouette, because rare medical conditions exist (e.g., adverse drug events, etc.) and should not be considered of less importance to frequently occurring ones.

\paragraph{Appropriate sampling} 
During our study of related work and existing implementations (\S\ref{sec:literature}), we observed that the only sampling strategy implemented was uniform. This strategy, however, is not appropriate for the macro-averaged Silhouette (\S\ref{ssec:sampling}). Therefore, we proposed a per-cluster sampling strategy, which we showed that it is considerably more robust compared to standard uniform sampling. This contribution can be particularly important for big datasets, because computing Silhouette is $\mathcal{O}(N^{2})$. As was shown in Figure~\ref{fig:sampling}, per-cluster sampling is robust to imbalance and yields approximately the same score even when the subsampled space is 2\% of the original (i.e., rightmost of Figure~\ref{fig:sampling}).

\section{Conclusions}
In this study, using a synthetic dataset, we show that the typical micro-averaged Silhouette Coefficient, which is often used to evaluate clustering solutions, is sensitive to cluster imbalance. By contrast, the macro-averaged Silhouette score, which is heavily overlooked in related literature, is far more robust. Therefore, although overlooked, we argue that macro-averaged Silhouette is a serious option that should be considered in cluster analysis. By studying macro-Silhouette further, we find that standard uniform sampling is not appropriate for this index, which is important because subsampling is an important step in the computation of Silhouette for large datasets. Hence, we proposed a novel robust per-cluster sampling strategy that follows in nature the macro-Silhouette computation. Finally, by employing eight real-world datasets of varying cluster-imbalance, we show that the Silhouette score and the estimated number of clusters differ in imbalanced datasets when using the two studied indices. 

\section*{Acknowledgements}
This work has been partially supported by project MIS 5154714 of the National Recovery and Resilience Plan Greece 2.0 funded by the European Union under the NextGenerationEU Program.

\bibliographystyle{unsrt}  
\bibliography{references}  

\appendix
\section{KMeans clustering for varying $k$}
\begin{figure}
    \centering
    \includegraphics[width=\textwidth]{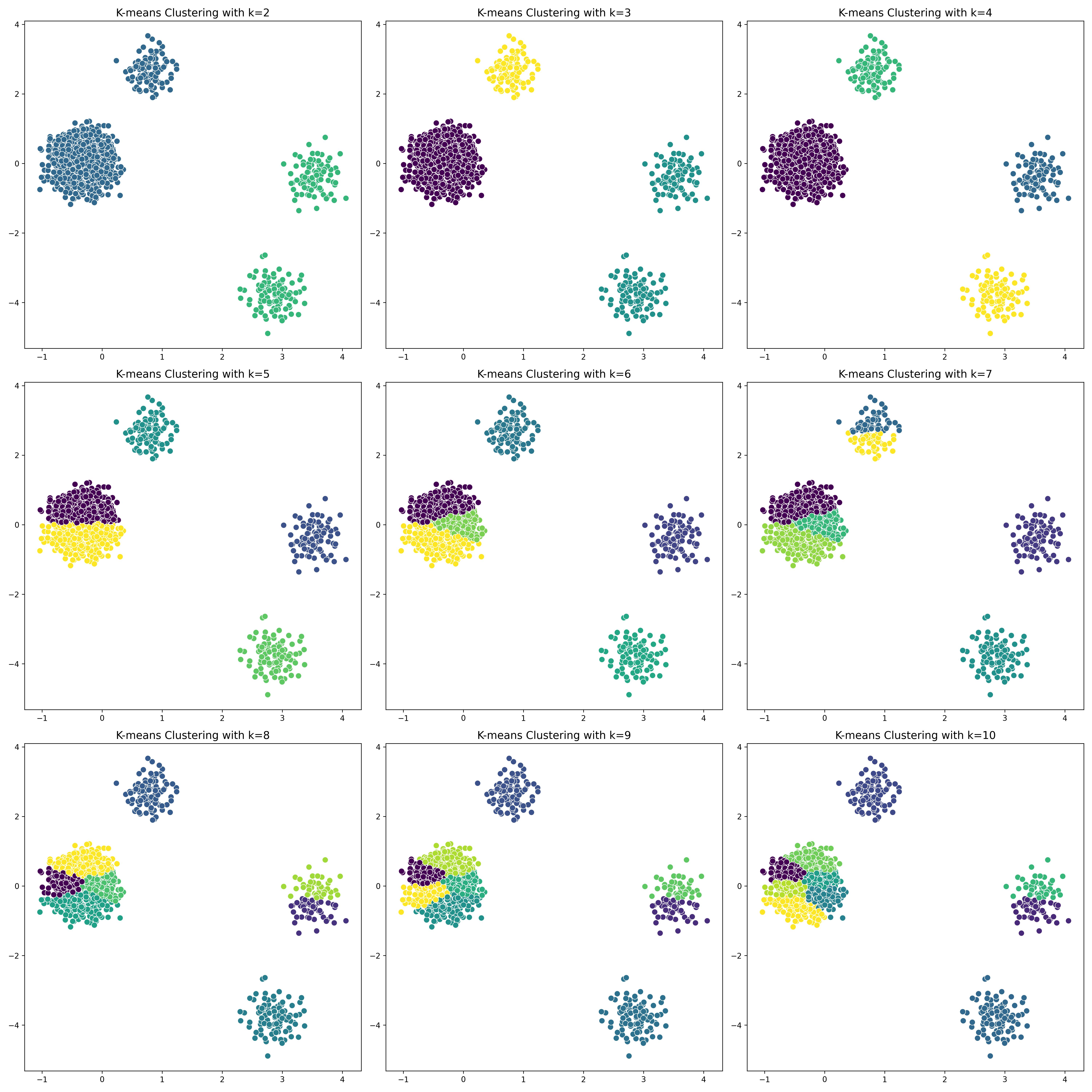}
    \caption{Clustering based on KMeans on a dataset of four isotropic Gaussian blobs, for varying number of clusters $k$. Colours reflect cluster assignments. For $k=4$, one colour is assigned per cluster.}
    \label{fig:findk_clusterings}
\end{figure}

Figure~\ref{fig:findk_clusterings} shows the clustering based on KMeans for a varying number of $k$. 

\end{document}